\documentclass[runningheads]{llncs}
\usepackage{graphicx}
\usepackage{algorithm}
\usepackage{algpseudocode}
\usepackage{amsmath}
\usepackage{array}
\usepackage[hidelinks]{hyperref}
\usepackage{caption}
\begin{document}
\title{TextGram: Towards a better domain-adaptive pretraining}

\titlerunning{TextGram: Towards a better domain-adaptive pretraining}

\author{Sharayu Hiwarkhedkar\inst{1,3} \and
Saloni Mittal \inst{1,3} \and
Vidula Magdum\inst{1,3} \and
Omkar Dhekane \inst{1,3} \and
Raviraj Joshi \inst{2,3} \and
Geetanjali Kale \inst{1} \and
Arnav Ladkat \inst{1,3}}
\authorrunning{Sharayu Hiwarkhedkar et al.}
%
\institute{Pune Institute of Computer Technology, Pune, Maharashtra India \and
 Indian Institute of Technology Madras, Chennai, Tamil Nadu, India \and
 L3Cube Labs, Pune}

\maketitle              
\begin{abstract}


For green AI, it is crucial to measure and reduce the carbon footprint emitted during the training of large language models. In NLP, performing pre-training on Transformer models requires significant computational resources. This pre-training involves using a large amount of text data to gain prior knowledge for performing downstream tasks. Thus, it is important that we select the correct data in the form of domain-specific data from this vast corpus to achieve optimum results aligned with our domain-specific tasks.
While training on large unsupervised data is expensive, it can be optimized by performing a data selection step before pretraining. Selecting important data reduces the space overhead and the substantial amount of time required to pre-train the model while maintaining constant accuracy. We investigate the existing selection strategies and propose our own domain-adaptive data selection method - \textit{TextGram} - that effectively selects essential data from large corpora.
We compare and evaluate the results of finetuned models for text classification task \textit{with} and \textit{without} data selection. We show that the proposed strategy works better compared to other selection methods.

\end{abstract}
\section{Introduction}

The pervasive adoption of the internet, interconnected devices, social media networks, and cloud-based services has led to the generation of an enormous volume of data on a global scale. The development of state-of-the-art NLP models hinges on the effective usage of this large-scale data to train transformer-based language models termed as PLMs (pre-trained language models). To meet the demands of training PLMs, data scientists and engineers often rely on high-end hardware in production environments, entailing significant time and computational resources. Consequently, a critical aspect of the preprocessing phase involves judiciously selecting and utilizing high-quality data to optimize the training time and compute requirements of these models.

Our proposed work aims to enhance the training efficiency of pretraining models by employing improved data selection techniques with various data metrics. We have explored and compared the most recent data selection training methods used for the optimal pretraining and finetuning tasks.

A pre-trained model is a deep neural network that has previously been trained on a large dataset and can be reused for a variety of downstream applications. Choosing how much data and which data samples to use to train a machine learning model is an important step in the field of Artificial Intelligence. If the data is insufficient, it may result in either overfitting (the model has low training errors but high test errors) or underfitting (the model is incapable of capturing the relationship in the data) situations. If the data is too large, it may entail a significant amount of computational time and resources. It could also result in negative transfer, which refers to a large amount of data having a negative impact on the model.

Moreover, we employ a 12-layered Bidirectional Encoder Representations from Transformers (BERT) model as the base model \cite{devlin2018bert}. BERT is trained on the complete English Wikipedia and Book corpus using Masked Language Modeling (MLM) as one of the pretraining strategies. In MLM, a portion of the input text is masked, and the model is trained to predict the masked words. This aids the model in comprehending word context and their relationships. Conducting adaptive pre-training with transformer models like BERT on out-of-domain corpus is a time and resource-intensive process.

A continuous pretraining of the BERT-based model with a large amount of general corpus data may take 3-4 days with a good GPU system and up to a week without GPU support. To shorten this duration, efficient data selection is necessary, wherein the dataset can be reduced while maintaining model accuracy. Data selection involves choosing the most relevant data from the vast corpus so that pretraining on the selected data can be accomplished in a shorter amount of time without compromising on results. Selecting the right data is thus a critical step. By implementing data selection before applying the pretraining task on huge datasets, we reduce the time and computational resources required. Through the exploration of available data selection techniques, we propose and implement a data selection strategy that enhances the pretraining task.


\section{Motivation}

Pre-training from scratch is a costly process with significant environmental impacts. Reducing the carbon footprint is a crucial factor to consider. A study at the University of Massachusetts reveals that the electricity consumed during transformer training can emit over 6,26,000 pounds of $CO_2$, which is five times more than a car's emissions \cite{parcollet2021energy}. By 2030, data centres might consume more than 6\% of the world's energy. Utilizing intelligent data selection techniques not only saves computational time and resources but also protects the environment by avoiding negative transfer and eliminating data with adverse impacts on the output.

In various survey papers, methods like N-Grams, IF-IDF, and Perplexity-based selections have been considered. However, some of these methods are computationally expensive, making them less suitable for production-level use. Developing a data selection method that improves pretraining efficiency with reduced computational cost will benefit NLP researchers in building more effective systems.

Our goal is to perform better research and provide superior solutions by developing efficient data selection techniques for pretraining.

\section{Related Work}
The previous work done, which focuses on data selection strategies, is discussed in this Section.
A few techniques that are majorly used in machine translation are context-dependent and independent-based functions. They include N-Grams, TF-IDF, Perplexity-based selection, Cross-entropy, etc. Generally, the subsets of data are varied for training. NMT iterates over the training corpus in several epochs, using a different subset in every iteration. Another method mentioned is gradual fine-tuning, where the training data is reduced in a few iterations \cite{van2017dynamic}.

Matthias et al. suggest a weighting scheme which allows us to sort sentences based on the frequency of unseen n-grams \cite{eck2005low}. The technique is applied to the BREC corpus with relatively simple sentences from the travel domain for the translation task. Marlies et al. use ngrams LMs to determine the domain relevance of sentence pairs and provide a comparative analysis with LSTM-based selection \cite{van2017dynamic}. The authors also introduce dynamic data selection - a method in which the selected subset of training data varies between different training epochs.
Additionally, Prafulla et al. use a TF-IDF-based approach for a clustering problem \cite{bafna2016document}. Another paper by Catarina et al. also proposes a variant of the TF-IDF data selection method \cite{cruz2018extracting}.

To compute the TF-IDF measure in their experiments, the authors first pre-process the corpus and consider every sentence in the domain corpus as a query sentence, and every sentence in the generic corpus as a document. Then, the authors obtain, for each query, a ranking of the documents computed with cosine similarity. This ranking is stored for every query sentence and used to retrieve the K-nearest neighbors (KNN) necessary to obtain different data selection sizes.

The work \cite{axelrod2011domain} investigates efficient domain adaptation for the task of statistical machine translation by retrieving the most pertinent sentences from a large general-domain parallel corpus. An assumption in domain adaptation, as described in this paper, is that a general-domain corpus of a sizable length will likely contain some sentences that are relevant to the target domain and should thus be used for training. The paper also highlights that the general-domain corpus is expected to contain sentences that are so dissimilar to the domain that using them to train our model will be more harmful than beneficial. It presents a method that applies language modeling techniques to Machine Translation, called the \textit{perplexity-based selection} method, which has been done by Gao et al. \cite{gao2002toward}, as well as by Moore and Lewis \cite{moore2010intelligent}. The ranking of the sentences in a general-domain corpus based on the in-domain's perplexity score has also been applied to machine translation by both Yasuda et al. \cite{yasuda2008method} and Foster et al. \cite{foster2010discriminative}.

Another perplexity-based method proposed by Moore and Lewis is \textit{cross entropy} and \textit{cross-entropy difference} \cite{moore2010intelligent}. A low perplexity sentence corresponds to a low cross-entropy sentence under the in-domain language model. The author used a language model trained on the Chinese side of the IWSLT corpus.

There are a few more complex techniques that are based on CNNs. Various algorithms are implemented under the data selection techniques like Recurrent Neural Network, EM-Clustering, Nearest Neighbor Selection, CNNs, a 5-parameter variation of FDA - FDA5, classification, etc.


\section{Experimentation Setup}
\label{exp-setup}
We discuss the experimental settings that we followed to obtain and evaluate efficient pre-training by data selection.


\subsection{\textbf{Datasets}}
For our experiment, we mainly use two kinds of data. One is \textit{out-domain}, which acts as a general domain corpus, and another is \textit{in-domain}, which is a domain-specific corpus. The in-domain data is smaller compared to the out-domain data. The summary of datasets used is shown in Table \ref{dataset_table}.


\begin{table*}[hbt!]
\centering
\setlength{\extrarowheight}{5pt}  
\begin{tabular}{|l|l|l|l|}
\hline
Dataset   & Description                        & Corpus Size        & Data Columns      \\ \hline
RealNews & This is a large corpus of news articles & 1M news     & Text, Score       \\ \hline
IMDb     & Movie review dataset               & 50,000
  reviews & Review sentiments \\ \hline
\end{tabular}
\vspace{5pt}
\caption{\label{dataset_table} Datasets used in the experiments}
\end{table*}

The \textit{RealNews} Dataset is a large corpus of news articles scraped from Common Crawl \cite{zellers2020defending}. It consists of 5000 news domains that are indexed by Google news. \textit{IMDb Movie Reviews} is a binary sentiment analysis dataset made up of 50,000 reviews from IMDb that have been classified as positive or negative \cite{maas-etal-2011-learning}. The IMDb Reviews dataset is widely used in many NLP subdomains. Here, we use the RealNews dataset as an out-domain dataset while IMDb Movie Reviews as an in-domain corpus. We evaluate our model on entertainment-specific domains.

\subsection{\textbf{Model Architecture}}
\begin{figure*}
    \centering 
    \fbox{\includegraphics[scale=0.38]{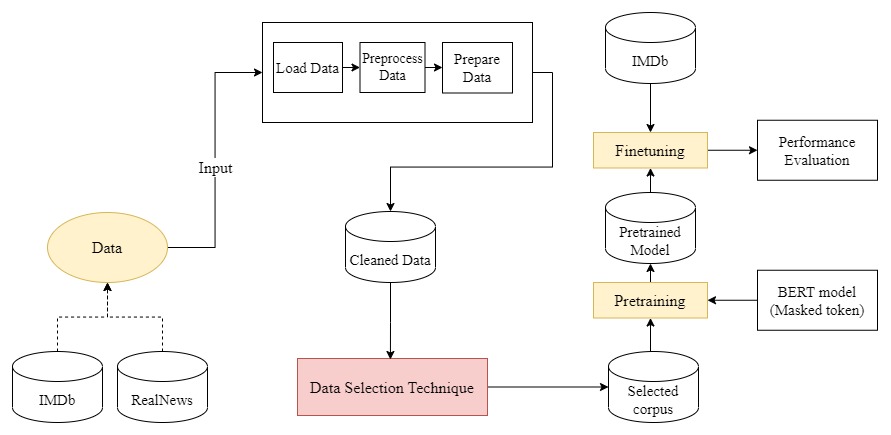}} 
    \caption{High-Level System Architecture Diagram - The corpus (both in-domain and out-domain) is first fed into pre-processing pipeline which will prepare the data for selection. Then, the selection strategy will be applied that will select data from out-domain based on in-domain training set. Further, selected corpus is used for continuous pre-training of BERT model. After pre-training, we perform fine-tuning to adapt the model on in-domain corpus.}
    \label{fig:architecture}
\end{figure*}

As shown in Fig. \ref{fig:architecture}, we collect and preprocess data and apply a selection strategy to obtain the best data. The selected data is then used for the second pretraining phase, with the objective of Masked Language Modeling (MLM). Finally, using the IMDb dataset \cite{ladkat2022towards}, we perform the classification task and analyze the results.

In the experiments, we utilize the Bidirectional Encoder Representations from Transformers (BERT) model for the MLM task and fine-tune it for text classification on the target dataset, IMDb dataset \cite{ladkat2022towards}. The BERT model comprises 12 layers of bidirectional transformer-based encoder blocks, where each layer consists of 12 self-attention heads. The BERT base cased model is pre-trained on a large English corpus using a self-supervised approach with the objective of Masked Language Modeling (MLM).

\subsection{\textbf{Data Selection Techniques}}

This section explains the data selection techniques used for experimentation. 
\subsubsection{\textbf{N-grams Coverage}}
N-grams define the sequence of 'n' contiguous elements from text or speech data. N-gram-based models are very popular in computational linguistics (for instance, statistical natural language processing). Many researchers have found this technique intuitive for exploring and implementing the variations of data selection techniques.

The notion is to select top\_k sentences from the general domain corpus based on the top n-grams from the target domain corpus. This is also called the adaptive approach as we consider the target domain for the selection method. A non-adaptive approach, on the other hand, only considers the general domain corpus and selects sentences based on the same notion of n-grams. \\
Selection Criteria (S):
 \begin{equation}
    S  = \bigl[ frequency(n-gram) \bigr]      
 \end{equation}

for n = 1,2,3... \\
Another selection approach based upon the n-gram model includes weighing each sentence in the corpus as a sum over the frequency of unseen n-grams, ranking the sentences based on the sentence weight, and then selecting top\_k sentences from the ranked list.
\\
\subsubsection{\textbf{TF-IDF based selection}}
This approach leverages the TF-IDF weighing scheme. TF-IDF measure is widely used in information retrieval. The term frequency (TF) defines the frequency of a term (t) in a document (d). The inverse document frequency (IDF) measures how much information a word can provide. Together, they are used to find the importance of a given word in a document.

The main idea of this approach is to represent each document D as a TF-IDF vector $(w_1,w_2,w_3,...w_m)$ where m is the size of the vocabulary. The TF-IDF weight ($w_k$) for the $k$th term in the term vector is calculated as,
 \begin{equation}
    w_k = TF_k * IDF_k  
 \end{equation}
where,
\begin{equation}
    TF_k = 
        \frac{\text{count of $k$ in $document (D)$}}{\text{no. of terms in $document (D)$}}
\end{equation}
and,
\begin{equation}
IDF_k = 
    \frac{\text{count of $documents$ in collection}}{\text{count of $documents$ containing $k$-th term}}
\end{equation}
\\
The similarity between two documents is calculated using cosine similarity and finally, documents are selected based upon the similarity score.
\\
\subsubsection{\textbf{Perplexity Based Data Selection}}
 In this method, the main idea is that the sentences in the general-domain corpus are scored by their perplexity score using an in-domain language model, and only the lowest ones are kept.

Perplexity is an intrinsic evaluation metric for language models. It entails determining some metric to evaluate the language model as a whole, without regard for the particular tasks it will be used for.

There are several approaches to perplexity-based data selection, the one used in this work is described below.

\paragraph{Perplexity as the normalized inverse probability of the test set}

Unigram-Normalized Perplexity as a Language Model Performance
Measure with Different Vocabulary Sizes \cite{roh2020unigram} explains the perplexity as the inverse probability of the test set, normalized by the number of words in the sequence. 

Models which assign probabilities to sequences of words $ ( w_1, w_2, w_3...w_n) $ are called language models. 

For a test set, 
\begin{equation}
    W =  w_1w_2w_3...w_n 
\end{equation}

By Chain Rule of Probability,
\begin{align}
P(W) &=  P(w_1)P(w_2|w_1)P(w_3|w_{1:2})...P(w_n|w_{1:n\ -1}) \\
P(W) &=  \Pi_{k=1}^n P(w_k|w_{1:{k\ -1}})         
\end{align}

An n-gram language model looks at the previous (n-1) words to estimate the next word in the sequence. For example, a bigram model will look at the previous 1 word, so:

\begin{equation}
    P(W) = P(w_{1:n}) = \Pi_{k=1}^n P(w_k|w_{k\ -1})
\end{equation}
 
 As per the definition of perplexity in this approach, the paper thus defines the formula for perplexity (PPL) for a Bigram model as,

\begin{equation} 
    PPL(W) = \sqrt[n]{P(w_{1:n})} =  \sqrt[n]{\Pi_{k=1}^n P(w_k|w_{k\ -1})}
\end{equation}

Therefore, as we're using the inverse probability, a lower perplexity indicates a better model. This method is dependent on the size of the sequence, and it also implies that adding more sentences to the dataset will introduce more uncertainty, thus reducing the probability and increasing the perplexity. To make the metric independent of the dataset size, normalization of the probability is done by dividing it by the total number of tokens to obtain a per-word measure. This is why we take the n-th root (where n is the length of the sequence) of the inverse probability of the sequence.
\\
\subsubsection{\textbf{Cross Entropy}}
Cross-entropy is a popular loss function in machine learning. It is used to compute the overall entropy between distributions. The loss function helps determine how effectively your program models the dataset.

In this method, we train a model on the entire dataset using cross entropy as the loss function. Next, we evaluate the model on the same dataset and calculate the cross entropy for each data point. The data points with low entropy scores are selected.

The cross-entropy score (CE) is calculated as follows:
\begin{equation}
    CE(S, p, q) = - \sum_{i=1}^{n} p(x_i)\log q(x_i) 
\end{equation}

Here, $x_i$ are the information units collected from a dataset and
p and q are the probability distribution over them;
p from the training dataset and q from the testing dataset.
\\
\subsubsection{\textbf{TextRank}}
TextRank is a graph-based language processing technique that is utilized for keyword and sentence extraction \cite{mihalcea2004textrank}. It is grounded on the PageRank algorithm, commonly used for ranking web pages. The input text is tokenized into text units (words or sentences) that represent the vertices/nodes of a graph. Two nodes are connected using the co-occurrence relation, meaning that nodes having N co-occurring units are connected with an edge. The score of a vertex is calculated using the following formula:


\begin{equation}
WS(V_i) = (1-d) + d \sum_{V_j \in In(V_i)} \frac{w_{j,i}}{\sum_{V_k \in Out(V_j)} w_{j,k}} WS(V_i)
\end{equation}

Here, $WS(V_i)$ represents the score of a vertex $V_i$. Set \textit{In} represents the set of vertices pointing to the vertex, and set \textit{Out} represents the set of vertices the vertex points to. $d$ is the damping factor, which describes the probability of visiting a vertex from the given vertex.

In the case of sentence extraction, similarity scores between sentences are calculated. Finally, the final vertex scores are sorted, and the top $n$ vertices are selected. TextRank is an unsupervised technique that does not require training on domain-specific corpora and can be effectively used for extractive summarization.
\\
\subsubsection{\textbf{Random Selection}}
Subsets of data (d) are collected at random from the huge dataset. The features or the type of data selected could not be promised in this case of data selection.

\section{Proposed technique - TextGram}
The existing TextRank technique has a few shortcomings, such as its non-adaptive nature, meaning it does not consider in-domain datasets during data selection. This lack of adaptiveness hampers the performance of the model when fine-tuned for downstream tasks. To address this limitation, we have introduced an n-gram technique in the initial processing phase of TextRank.

In our proposed approach, we start by selecting the top $k$ in-domain sentences based on the highest frequencies of n-grams calculated from the in-domain corpus. These selected in-domain sentences are then combined with the out-domain corpus. Subsequently, we perform paraphrase mining on this combined set of sentences to determine similarity scores for various pairs of sentences.

To construct a graph representation, we create a sparse matrix using the similarity scores, where the nodes represent sentences and the edges indicate the scores between the sentences. This graph is then fed into the PageRank algorithm, which computes the scores of each sentence based on the node weights. The obtained scores are then sorted in descending order, and we select the top N sentences.

\begin{figure*}[ht!]
    \centering 
    \fbox{\includegraphics[width = \textwidth]{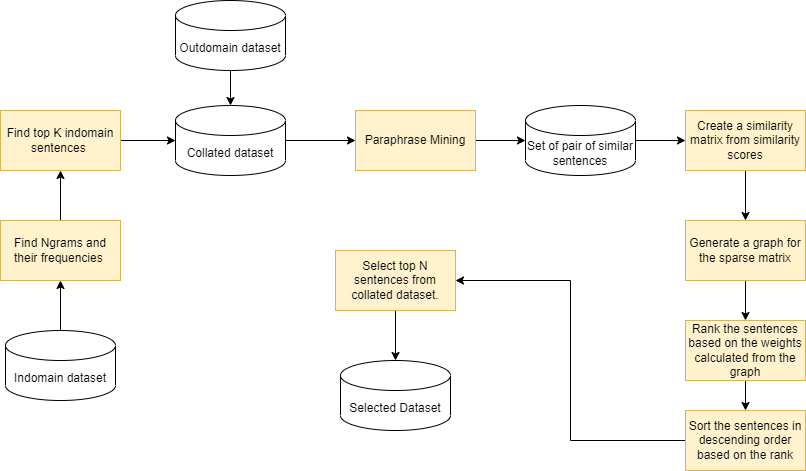}}
    \caption{Architecture diagram: TextGram based ranking}
    \label{fig:textgram}
\end{figure*}

Next, we separate the in-domain sentences from the selected sentences, resulting in the final selected out-domain dataset. As a result of incorporating n-grams, our technique yields better results compared to the original TextRank approach. Refer to Fig. \ref{fig:textgram} for the TextGram architecture and Algorithm \ref{alg:three} for the brief algorithm.\\

\begin{algorithm}
\caption{TextGram}\label{alg:three}
\vspace{0.1cm}
\begin{algorithmic}
\label{alg:MYALG}
 \renewcommand{\algorithmicrequire}{\textbf{Input:}}
 \renewcommand{\algorithmicensure}{\textbf{Output:}}
\Require{IMDb dataset (50K), Realnews (1M)}
\Ensure{selectedRealnews dataset (0.25M)}
 \State IMDb $\gets$ IMDb dataset;
 \State bigrams $\gets $ \{\};

 \For{sentence in IMDb}
  \State tokens $\gets $ wordTokenize(sentence);
  \For{i in range (len(tokens))}
    \State bigram $\gets$ tokens[i] + tokens[i+1];
    \State bigrams[bigram] $\gets$ bigrams[bigram]+1;
  \EndFor
  \EndFor
  \State Sort $bigrams$ dictionary in descending order based on frequency values.
  \State TopBiGrams $\gets bigrams[0...100]$
  \State selectedIMDb $\gets$ []
  \For{sentence in IMDb}
    \For{bigram in TopBiGrams}
        \If{bigram in sentence}
            \State selectedIMDb $\gets$ sentence;
        \EndIf
    \EndFor
    \EndFor
  \State realnews $\gets$ Realnews;
  \State collatedDataset $\gets$ [];
  
  \For{sentence in realnews}
    \State collatedDataset.append(\($sentence$, $'Realnews'$)\)
    \EndFor
    
    \For{sentence in selectedIMDb}
    \State collatedDataset.append(\($sentence$, $'IMDb'$)\)
    \EndFor

  \State SimilarityScores $\gets$ paraphraseMining(collatedDataset);  // Similarity score is a list of 3-tuple element: //$($score$, $sentence1Index$, $sentence2Index$)$
  \State similarityMatrix $\gets$ [][]
  \For{i in range(len(SimilarityScores)}
  \State sentence1Index = SimilarityScores[i][1]
  \State sentence2Index = [SimilarityScores[i][2]
  \State similarityMatrix[sentence1Index][sentence2Index] $\gets$ SimilarityScores[i][0]
   \State similarityMatrix[sentence2Index][sentence1Index] $\gets$ SimilarityScores[i][0]
   \EndFor
   \State sentencesGraph $\gets$ Graph(similarityMatrix)
   \State rankings $\gets$ = Pagerank(sentencesGraph)
   \State Sort $rankings$ list of tuples in descending order based on rank of sentences.
  \State selectedRealnews $\gets$ []
  \For{i in range(2500000)}
        \State tupleSentence $\gets$ collatedDataset[ranking[i][0]]
        \If{tupleSentence[1] $=$ 'Realnews'}
            \State selectedRealnews $\gets$ tupleSentence[0]
 ;
        \EndIf
    \EndFor
    

\end{algorithmic}
\end{algorithm}





\section{\textbf{Evaluation Results}}
\subsection{\textbf{Fine-tuning without Data Selection}}
Below we discuss the results of our classification experiments using  \textit{with selection} and \textit{without selection} strategy and measure the classification quality using performance metrics such as Accuracy, Precision, Recall, and F1-score.

TABLE \ref{without_selection_table} shows evaluation scores for our classification task. We got the following results after performing the pre-training step on \textit{RealNews} examples \textbf{Without Data Selection} and then performing the downstream task with the IMDb dataset.  

\begin{table}
\centering

\setlength{\extrarowheight}{3pt}  
\begin{tabular}{ |l|l|l|l| }
  \hline
  \textbf{Accuracy (\%)} & \textbf{Recall (\%)} & \textbf{Precision (\%)} & \textbf{F1 Score (\%)} \\ 
  \hline
   90.40 & 89.91 & 90.80 & 90.36
 \\ 
  \hline
\end{tabular}
\vspace{5pt}
\caption{\label{without_selection_table} Fine-tuning results without data selection}
\end{table}
The above table shows that, When we directly fine-tune our RealNews pre-trained model using the IMDb review dataset with no selection strategy applied, we see a very small change in F1-score value compared to F1-scores using \textit{with-selection} (see TABLE \ref{with_selection_table}).


\subsection{\textbf{Fine-tuning with Data Selection}}
\begin{table}[ht!]
\resizebox{\columnwidth}{!}{
\setlength{\extrarowheight}{3pt}  
\begin{tabular}{ |l|l|l|l|l| }
  \hline
  \textbf{Data Selection Technique} & \textbf{Accuracy (\%)} & \textbf{Recall (\%)} & \textbf{Precision (\%)} & \textbf{F1 Score (\%)} \\
  \hline
  Random Selection & 90.77 & 90.32 & 91.14 & 90.73  \\ 
  \hline
  N-Grams & 91.03	& 90.44	& 91.52	& 90.97 \\ 
  \hline
  Perplexity & 90.99	& 90.96 & 91.02 &	90.99 \\ 
  \hline
  Cross-Entropy & 90.81 & 90.42 & 91.14 & 90.77 \\ 
  \hline
  TextRank & 90.60 & 89.23 & 91.75	& 90.47\\ 
  \hline
  \textbf{TextGram} & \textbf{91.02} & \textbf{91.02} & \textbf{91.02} & \textbf{91.02} \\
  \hline
\end{tabular}
}
\vspace{5pt}
\caption{\label{with_selection_table} 
The classification results after fine-tuning of bert-base model on IMDb dataset. Before fine-tuning, the bert-base is first pretrained on the out-domain selected data. Above table shows the effect of various data selection strategies on the performance of the model.}
\end{table}

The TABLE \ref{with_selection_table} shows a comparative analysis of fine-tuning results on the IMDb dataset, using various data selection strategies which are discussed in section \ref{exp-setup}. Our evaluation shows that the TextGram-based ranking strategy outperforms the other selection strategies.

\section{Conclusion and Future Work}
An extensive literature review of the most widely used methods for effective data selection tasks was conducted for pretraining. We implemented these techniques and developed a suitable and optimum technique from the same. The scores say that N-Grams gave the highest accuracy. Also, TextRank is studied to have better impacts when combined with N-Grams because of its graph-based complex but deep implementations and research. TextRank is mostly used for sentence similarity and keyword extraction, but by studying the Page Rank algorithm, we developed this technique to work on domain adaptive data selection tasks. The techniques implemented give a score of around 1\% similar to or higher than the baseline numbers computed without data selection.

\section*{Acknowledgments}
This work was done under the L3Cube Pune mentorship
program. We would like to express our gratitude towards
our mentors at L3Cube for their continuous support and
encouragement.

\bibliographystyle{splncs04}
\bibliography{temp}

\begin{thebibliography}{10}
\providecommand{\url}[1]{\texttt{#1}}
\providecommand{\urlprefix}{URL }
\providecommand{\doi}[1]{https://doi.org/#1}

\bibitem{axelrod2011domain}
Axelrod, A., He, X., Gao, J.: Domain adaptation via pseudo in-domain data
  selection. In: Proceedings of the 2011 conference on empirical methods in
  natural language processing. pp. 355--362 (2011)

\bibitem{bafna2016document}
Bafna, P., Pramod, D., Vaidya, A.: Document clustering: Tf-idf approach. In:
  2016 International Conference on Electrical, Electronics, and Optimization
  Techniques (ICEEOT). pp. 61--66. IEEE (2016)

\bibitem{cruz2018extracting}
Cruz~Silva, C., Liu, C.H., Poncelas, A., Way, A.: Extracting in-domain training
  corpora for neural machine translation using data selection methods.
  Association for Computational Linguistics (2018)

\bibitem{devlin2018bert}
Devlin, J., Chang, M.W., Lee, K., Toutanova, K.: Bert: Pre-training of deep
  bidirectional transformers for language understanding. arXiv preprint
  arXiv:1810.04805  (2018)

\bibitem{eck2005low}
Eck, M., Vogel, S., Waibel, A.: Low cost portability for statistical machine
  translation based on n-gram frequency and tf-idf. In: Proceedings of the
  Second International Workshop on Spoken Language Translation (2005)

\bibitem{foster2010discriminative}
Foster, G., Goutte, C., Kuhn, R.: Discriminative instance weighting for domain
  adaptation in statistical machine translation. In: Proceedings of the 2010
  conference on empirical methods in natural language processing. pp. 451--459
  (2010)

\bibitem{gao2002toward}
Gao, J., Goodman, J., Li, M., Lee, K.F.: Toward a unified approach to
  statistical language modeling for chinese. ACM Transactions on Asian Language
  Information Processing (TALIP)  \textbf{1}(1),  3--33 (2002)

\bibitem{ladkat2022towards}
Ladkat, A., Miyajiwala, A., Jagadale, S., Kulkarni, R., Joshi, R.: Towards
  simple and efficient task-adaptive pre-training for text classification.
  arXiv preprint arXiv:2209.12943  (2022)

\bibitem{maas-etal-2011-learning}
Maas, A.L., Daly, R.E., Pham, P.T., Huang, D., Ng, A.Y., Potts, C.: Learning
  word vectors for sentiment analysis. In: Proceedings of the 49th Annual
  Meeting of the Association for Computational Linguistics: Human Language
  Technologies. pp. 142--150. Association for Computational Linguistics,
  Portland, Oregon, USA (Jun 2011), \url{https://aclanthology.org/P11-1015}

\bibitem{mihalcea2004textrank}
Mihalcea, R., Tarau, P.: Textrank: Bringing order into text. In: Proceedings of
  the 2004 conference on empirical methods in natural language processing. pp.
  404--411 (2004)

\bibitem{moore2010intelligent}
Moore, R.C., Lewis, W.: Intelligent selection of language model training data.
  In: Proceedings of the ACL 2010 conference short papers. pp. 220--224 (2010)

\bibitem{parcollet2021energy}
Parcollet, T., Ravanelli, M.: The energy and carbon footprint of training
  end-to-end speech recognizers  (2021)

\bibitem{roh2020unigram}
Roh, J., Oh, S.H., Lee, S.Y.: Unigram-normalized perplexity as a language model
  performance measure with different vocabulary sizes. arXiv preprint
  arXiv:2011.13220  (2020)

\bibitem{van2017dynamic}
Van Der~Wees, M., Bisazza, A., Monz, C.: Dynamic data selection for neural
  machine translation. arXiv preprint arXiv:1708.00712  (2017)

\bibitem{yasuda2008method}
Yasuda, K., Zhang, R., Yamamoto, H., Sumita, E.: Method of selecting training
  data to build a compact and efficient translation model. In: Proceedings of
  the Third International Joint Conference on Natural Language Processing:
  Volume-II (2008)

\bibitem{zellers2020defending}
Zellers, R., Holtzman, A., Rashkin, H., Bisk, Y., Farhadi, A., Roesner, F.,
  Choi, Y.: Defending against neural fake news (2020)

\end{thebibliography}
\end{document}